\documentclass{llncs}
\usepackage{graphicx}
\usepackage{comment}
\usepackage{amsmath,amssymb} 
\usepackage{color}

\usepackage[width=122mm,left=12mm,paperwidth=146mm,height=193mm,top=12mm,paperheight=217mm]{geometry}

\begin{document}
\pagestyle{headings}
\mainmatter
\def\ECCVSubNumber{4775}  

\title{Semantics-Driven Unsupervised Learning for Monocular Depth and Ego-Motion Estimation} 



\titlerunning{}
%
\author{Xiaobin Wei \and
Jianjiang Feng \and
Jie Zhou}

\institute{}
%

\maketitle

\begin{abstract}
  We propose a semantics-driven unsupervised learning approach for monocular depth and ego-motion 
  estimation from videos in this paper. Recent unsupervised learning methods 
  employ photometric errors between synthetic view and actual image as a supervision signal for training. 
  In our method, we exploit semantic segmentation information to mitigate the effects of dynamic objects and occlusions in the scene, 
  and to improve depth prediction performance by considering the correlation between depth and semantics.
  To avoid costly labeling process, we use noisy semantic segmentation results obtained by a pre-trained semantic segmentation network.
  In addition, we minimize the position error between the corresponding points of adjacent frames 
  to utilize 3D spatial information. Experimental results on the KITTI dataset show that 
  our method achieves good performance in both depth and ego-motion estimation tasks.
\keywords{Depth prediction, ego-motion estimation, semantics-driven unsupervised learning}
\end{abstract}

\section{Introduction}
Visual odometry (VO) \cite{scaramuzza2011visual} is a process of estimating camera's motion by 
using image sequences as input. It is a basic task in many computer vision 
applications such as automatic driving, augmented reality, navigation systems, etc. 
Recovering 3D depth information from 2D images is an important problem in computer 
vision, which plays an important role in scene understanding \cite{zhang2017physically,sakaridis2018model}, 
3D reconstruction \cite{schoenberger2016sfm,sweeney2015optimizing}, etc. 

In the past decade, geometry-based VO approaches have been extensively studied. 
There are usually two kinds of methods: (1) Feature-based methods, such as 
PTAM \cite{klein2007parallel} and ORB-SLAM \cite{mur2015orb,mur2017orb}, 
estimate camera poses and generate sparse 3D map by minimizing the re-projection error. 
Typical processes include feature extraction, feature matching, motion 
estimation and local optimization, estimating camera pose and generating 
sparse 3D maps. (2) Direct methods, according to raw pixel intensity of 
images directly calculate the camera motion by minimizing the photometric 
error. However, these methods are often not robust enough in challenging 
environments, such as motion blurring and lack of texture. 
In recent years, some research works \cite{eigen2014depth,eigen2015predicting,liu2015learning,laina2016deeper,li2018undeepvo,ummenhofer2017demon} 
have adopted supervised neural networks to solve the VO and depth prediction 
problems. Since these methods need a lot of labeled data with ground truth to train, 
and LIDAR sensors in autonomous vehicles provide only very sparse 3D points, 
their generalization to new scenarios is limited.

Compared with supervised learning, unsupervised learning does not require labeled data. 
More and more studies focus on unsupervised learning methods for depth and camera motion 
estimation \cite{zhou2017unsupervised,li2018undeepvo,zhan2018unsupervised,mahjourian2018unsupervised,yin2018geonet,yang2018lego,shen2019icra,luo2018pixel,ranjan2019competitive,wang2019recurrent,godard2019digging}. 
They use photometric error as a loss function to learn. Depth and poses are 
used to project the source image onto the target frame for synthesizing the target 
view, and the network is trained by minimizing the error between the synthesized 
view and the actual image.

For dynamic objects and occluded objects in the scene, however, the assumption of photometric consistence between adjacent frames does not hold, 
which will lead to inaccurate depth prediction. 
We propose to use semantic segmentation to alleviate this problem. 
When the labels of the pixel in the source image and the corresponding pixel in the target image 
are different, it may be a moving object or occlusion. 
We further explore another application of semantic segmentation by utilizing the correlation between depth and semantics to improve depth estimation performance. For example, 
between the adjacent upper and lower pixels in the ground area of the image, the upper pixel has a larger depth. 
To avoid the costly manual labeling process, we use semantic segmentation obtained from a pre-trained segmentation network. 

Although the semantic segmentation algorithm requires supervised learning, 
its adoption does not affect the unsupervised nature of the core algorithm in this paper, 
and it is reasonable in practice, because of the following reasons: 
(1) it is not necessary to label the semantics on the training data of the depth prediction and pose estimation network; 
(2) Obtaining semantic segmentation labels requires labor costs, but it is feasible. In contrast, there is currently no convenient technology to easily obtain high-resolution, 
accurate depth maps in dynamic scenes; 
(3) Semantic segmentation is essentially supervised; 
(4) There are already multiple large-scale labeled semantic segmentation datasets; 
(5) Existing semantic segmentation networks have achieved excellent performance and generalization ability.

\begin{figure}[t]
  \begin{center}
  \includegraphics[width=0.6\linewidth]{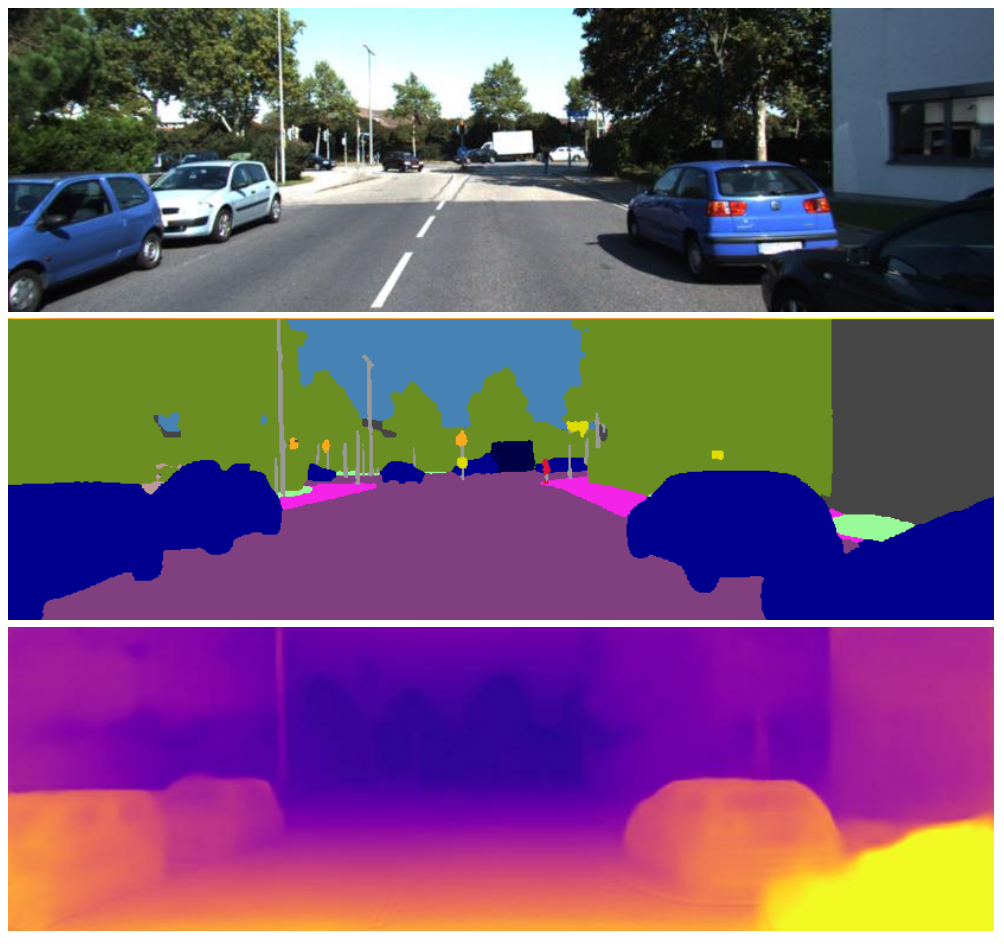}
  \end{center}
     \caption{Example of depth prediction on the KITTI dataset. Top to bottom: input RGB image, input semantic segmentation estimated by \cite{zhu2019improving}, and outputted depth map by the proposed method}
  \label{fig:fig1}
  \end{figure}

Photometric error considers only 2D appearance information. 
We propose an additional 3D point loss, which considers 3D spatial information. 
For a pixel in the target frame, its 3D coordinate in the target frame coordinate system 
can be obtained from the depth map. 
The 3D coordinate of the corresponding pixel in the source frame coordinate system can also be obtained by the depth maps and the transformation matrix.
According to the transformation matrix, they can be transformed into a same 
coordinate system, and the distance between the two points should be as close 
as possible, which can be used as the 3D point loss.

Our method is evaluated on KITTI dataset \cite{geiger2012we}, and the results show the 
effectiveness of our method in monocular depth prediction and camera motion estimation. Figure~\ref{fig:fig1} shows the result of our monocular depth prediction on 
the KITTI dataset. Our main contributions are as follows: (1) We propose a 
semantic loss and using semantic consistency as a mask for photometric loss and 3D point loss to reduce the influence of dynamic objects and occlusion in the scene, and using the depth characteristics of certain semantic category to improve 
the accuracy of depth prediction. 
(2) We propose a 3D point loss to improve the performance of depth prediction by utilizing 3D information. 

\section{Related Work}
\label{sec:related}
Existing methods for depth and self-motion estimation include geometry-based 
methods and learning-based methods.

{\bf Geometry-based methods.} Geometry-based VO schemes can be divided into two 
categories: feature-based methods and direct methods. Feature-based methods 
extract stable feature points from each frame, complete the matching of adjacent 
frames through the invariant descriptors \cite{lowe2004distinctive,bay2006surf,rublee2011orb} of these feature points, and then 
recover camera poses and map point coordinates more robustly through 
epipolar geometry \cite{hartley2003multiple}, but extraction and matching of feature points is 
time-consuming, which makes the classical feature-based methods run slower 
than direct methods. MonoSLAM \cite{davison2007monoslam} proposed by 
Davison et al. in 2007 is the first real-time monocular visual SLAM system. 
The front-end uses feature points tracking method and the back-end uses extended 
Kalman filter technology. Klein et al. proposed PTAM (Parallel tracking and mapping) 
\cite{klein2007parallel}, which was the earliest method to use non-linear optimization, 
and implemented the parallelization of tracking and mapping processes. 
Mur-Artal et al. proposed ORB-SLAM \cite{mur2015orb} in 2015, which is based on PTAM 
architecture, adds map initialization and loop closure detection, optimizes methods 
of key frame selection and map construction, and achieves good results in processing 
speed, tracking effect and map accuracy. Direct methods directly estimate the 
camera poses and map structure through minimizing photometric error without 
calculating key points and descriptors. LSD-SLAM (Large-scale direct monocular SLAM) 
\cite{engel2014lsd}, which is a monocular SLAM algorithm based on direct method proposed by 
Engel et al. in 2014, uses direct tracking method and is insensitive to the 
missing homogenous regions. 
Engel et al. \cite{engel2017direct} combine a fully direct probabilistic model with consistent, 
joint optimization of all model parameters, including geometry-represented to estimate 
camera internal parameters, pose and depth of pixels.

{\bf Supervised learning methods.} Eigen et al. \cite{eigen2014depth} propose a 
multi-scale deep network to solve the depth prediction problem. They address 
the problem by employing two deep network stacks: one that makes a coarse global prediction 
based on the entire image, and another that refines this prediction locally. 
\cite{eigen2015predicting} is an extension of \cite{eigen2014depth}, which solves
three different computer vision tasks using a single multiscale convolutional 
network architecture: depth prediction, surface normal estimation, and semantic labeling. 
Liu et al. \cite{liu2015learning} consider depth estimation as a continuous CRF learning 
problem, which learns the unary and pairwise potentials of continuous CRF in a unified 
deep CNN framework. Laina et al. \cite{laina2016deeper} propose a fully convolutional 
architecture, encompassing residual learning, to model the ambiguous mapping between 
monocular images and depth maps. \cite{xie2016deep3d,ranftl2016dense,kendall2017end} use more than one image during training stage for depth estimation.
\cite{kuznietsov2017semi} is a semi-supervised method to learn dense monocular depth. 
In terms of VO, Wang et al. \cite{wang2017deepvo} 
proposed an end-to-end monocular framework, DeepVO, which not only automatically 
learns effective feature representation for the VO problem through CNN, 
but also implicitly models sequential dynamics and relations using deep RNN. 
Ummenhofer et al. \cite{ummenhofer2017demon} use a convolutional network consisting 
of multiple stacked encoder-decoder networks for end-to-end training to compute 
depth and camera motion from successive, unconstrained image pairs. VINet \cite{clark2017vinet} 
is a sequence-to-sequence framework for motion estimation using visual and inertial sensors.
\cite{kendall2015posenet,kendall2017geometric,walch2017image,brahmbhatt2018geometry} regress the 6-DOF camera pose from a single RGB image.

{\bf Unsupervised learning methods.} Most of the unsupervised works are supervised by 
view synthesis, which minimizes the difference between the synthesized view and the 
target image. Godard et al. \cite{godard2017unsupervised} propose a novel training 
loss that enforces consistency between the disparities produced relative to both 
the left and right images to estimate depth. Zhou et al. \cite{zhou2017unsupervised} 
propose an unsupervised learning framework for the task of monocular depth and 
camera motion estimation from unstructured video sequences. The network is divided 
into two parts: single-view depth and multi view pose networks, with a loss based 
on warping nearby views to the target using the computed depth and poses. 
UnDeepVO \cite{li2018undeepvo} makes use of spatial losses and temporal losses 
between stereo image sequences for unsupervised training, for which they use stereo 
image pairs to recover the scale but test it by using consecutive monocular images. 
Zhan et al. \cite{zhan2018unsupervised} add deep feature-based warping loss to 
loss function to improve the accuracy and robustness of depth and motion estimation. 
Considering the inferred 3D geometry of the whole scene, Mahjourian et al. 
\cite{mahjourian2018unsupervised} proposed an unsupervised learning method for 
monocular image depth and motion estimation using 3D geometric constraints to 
enforce consistency of the estimated 3D point clouds and ego-motion across consecutive frames. 
GeoNet \cite{yin2018geonet} is a jointly unsupervised learning framework for monocular 
depth, optical flow and ego-motion estimation from videos, which uses separate components 
to learn the rigid flow and object motion by rigid structure reconstructor and non-rigid motion 
localizer respectively. Some works \cite{rezende2016unsupervised,yan2016perspective,tatarchenko2016multi} 
learn 3D structures from 2D images based on the projective geometry. 
Wang et al. \cite{wang2018learning} using a differentiable implementation of direct visual odometry and a novel depth normalization strategy 
to improve monocular video depth prediction. 
Yang et al. \cite{yang2018lego} introduce a “3D as-smooth-as-possible (3D-ASAP)” prior to learn edges and geometry (depth, normal) all at once. 
Shen et al. \cite{shen2019icra} use epipolar geometry to incorporate intermediate geometric computations such as feature matches into the tasks. 
In order to eliminate the need of static scene assumption, three parallel networks are used to predict the camera motion, depth map, and per-pixel optical flow between two frames in EPC++ \cite{luo2018pixel}. 
Ranjan et al. \cite{ranjan2019competitive} introduce Competitive Collaboration to segment the scene into static and moving regions without supervision. 
Wang et al. \cite{wang2019recurrent} use Recurrent Neural Networks to utilize the temporal information. 
Chen et al. \cite{Chen_2019_CVPR} do unsupervised depth prediction and supervised semantic segmentation using stereo images pairs and semantic segmentation ground truth. 
Meng et al. \cite{Meng_2019_CVPR} use semantic segmentation, instance class segmentation and instance edge map for unsupervised 3D geometry perception. 
Godard et al. \cite{godard2019digging} propose a minimum reprojection loss to robustly handle occlusions and a full-resolution multi-scale sampling method to reduce visual artifacts. 
We take semantic consistency as the mask of photometric loss and 3D point loss, 
and consider the correlation between depth and semantics. 

\begin{figure*}[t]
  \begin{center}
  \includegraphics[width=1\linewidth]{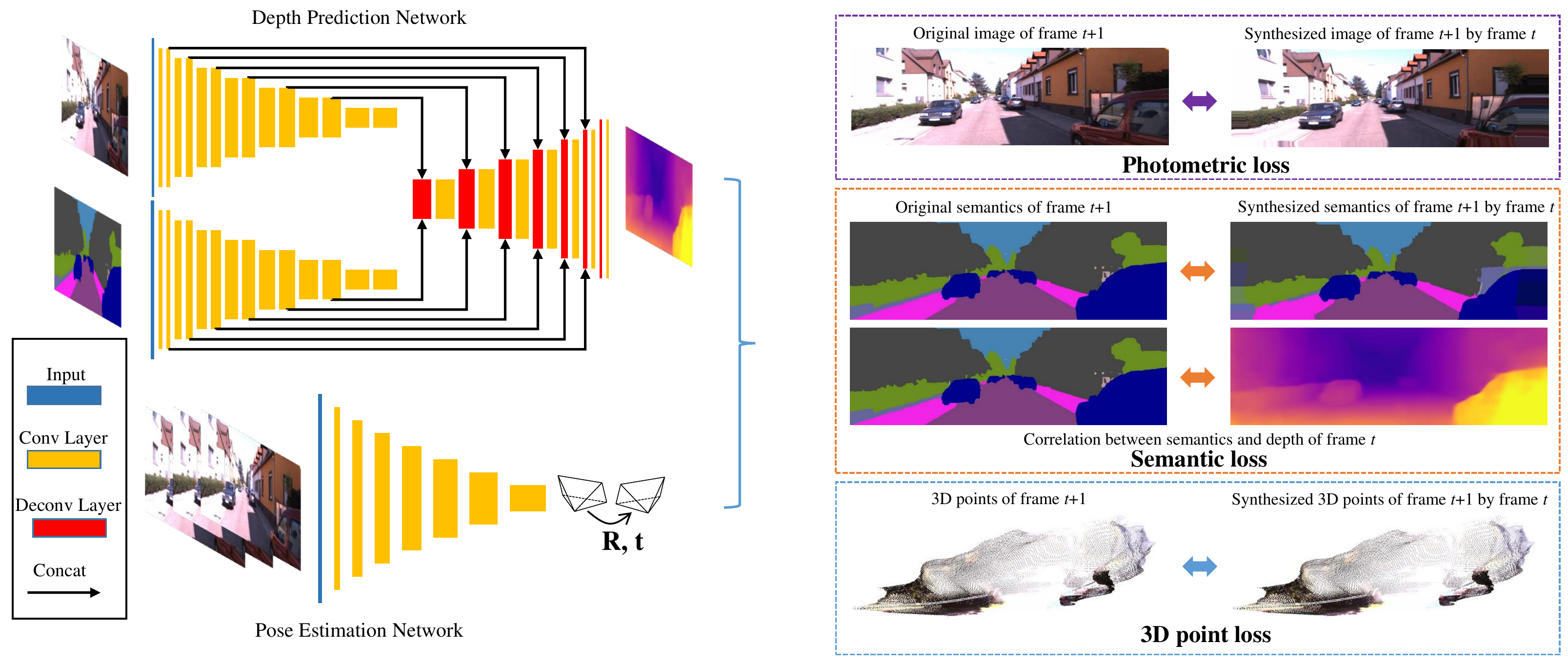}
  \end{center}
     \caption{Architecture of our method. The network consists of two parts: depth prediction 
     network and pose estimation network, which are trained jointly. Given as input RGB video 
     and semantic labels estimated by a state-of-the-art semantic segmentation algorithm, 
     the proposed network outputs depth map of each frame and relative pose between adjacent frames}
  \label{fig:fig2}
  \end{figure*}

\section{Method}
\label{sec:method}
An overview of our approach is shown in Figure~\ref{fig:fig2}. It can learn depth and camera 
motion from unlabeled data. The network consists of two parts: depth prediction network and 
pose estimation network, which are trained jointly. The framework takes as input a sequence 
of consecutive monocular images and semantic segmentation. The depth prediction network outputs the depth map of each frame, and the pose 
estimation network outputs pose between adjacent frames. The loss function includes photometric loss, semantic loss 
and 3D point loss.

\subsection{Photometric Loss}
\label{sec:2dloss}

In previous methods, image reconstruction loss is used, which is a fundamental supervision signal widely used in 
unsupervised tasks. For two adjacent frames, $I_t$ and $I_{t'}$, if the depth map of $I_t$ 
and the relative pose between the two views are given, then $I_t$ view can be reconstructed 
from $I_{t'}$. Taking $I_t$ as input, depth prediction network generates depth map for $I_t$, 
denoted as $\hat{D}_t$. The relative camera pose between two views can be estimated from the 
pose estimation network, denoted as $\hat{T}_{t \to t'}$. Denote $p_t$ as the homogeneous 
coordinates of a pixel in $I_t$, and $p_{t'}$ as the corresponding pixel in $I_{t'}$. 
Using epipolar geometry, the projected coordinates can be expressed as: 
\begin{equation}   
p_{t'} \sim K\hat{T}_{t \to t'}\hat{D}_t(p_t)K^{-1}p_t
\label{equ:projected}
\end{equation}
where $K$ is the camera intrinsic matrix, $\hat{T}_{t \to t'}$ is the camera coordinate 
transformation matrix from the $I_t$ frame to the $I_{t'}$ frame, $\hat{D}_t(p_t)$is the 
depth value of the $p_t$ pixel in the $I_t$ frame, and the coordinates are homogeneous.

According to the projection relationship, a new synthetic frame $\hat{I}_{t' \to t}$ can be obtained 
from $I_{t'}$ frame by using the differentiable bilinear interpolation mechanism proposed 
in \cite{jaderberg2015spatial}.

Structural similarity (SSIM) \cite{wang2004image} can be used to evaluate the quality of image prediction. 
A widely used image reconstruction error function is as follows: 
\begin{equation}   
  re(I_t^{u,v}, \hat{I}_{t' \to t}^{u,v})= \frac{\alpha}{2}(1-SSIM(I_t^{u,v}, \hat{I}_{t' \to t}^{u,v}))  + (1-\alpha)\Vert I_t^{u,v}-\hat{I}_{t' \to t}^{u,v} \Vert _1
\label{equ:reconerr}
\end{equation}
where the superscript $uv$ represents the image pixel at coordinates $(u, v)$ and $\alpha$ usually is set to 0.85.

The image reconstruction loss can be formulated as:
\begin{equation}   
   L_{recon}=\sum_{u,v}re(I_t^{u,v}, \hat{I}_{t' \to t}^{u,v}).
\label{equ:reconloss}
\end{equation}

\cite{godard2019digging} proposes a minimum reprojection loss to handle occlusions. \cite{godard2019digging} applies a per-pixel mask $\mu$:
\begin{equation}   
  \mu^{u,v}=\left[ \min_{t'}re(I_t^{u,v}, \hat{I}_{t' \to t}^{u,v}) < \min_{t'}re(I_t^{u,v}, I_{t'}^{u,v}) \right]
\label{equ:mu}
\end{equation}
where $t' \in \{ t-1, t+1\}$ and $\left[ \cdot \right]$is the Iverson bracket. The minimum reprojection loss is:
\begin{equation}   
  L_{p}=\sum_{u,v}\mu^{u,v}\min_{t'}re(I_t^{u,v}, \hat{I}_{t' \to t}^{u,v}).
\label{equ:minloss}
\end{equation}

In order to solve the gradient-locality issue in motion estimation and eliminate the 
discontinuity of the depth learned in low texture regions, depth smoothness loss is 
used to adjust the depth estimation. We adopt the depth gradient smoothness loss in 
\cite{godard2017unsupervised} which uses image gradient to weight depth gradient:
\begin{equation}   
   L_{smooth}=\sum_{u,v}\vert \nabla D_t^{u,v} \vert^T \cdot e^{-\vert \nabla I_t^{u,v} \vert}
\end{equation}
where $\nabla$ is the vector differential operator, $T$ denotes the transpose of image 
gradient weighting and $\vert \cdot \vert$ denotes elementwise absolute value.

\subsection{Semantic Loss}
\label{sec:ssloss}
Similar to photometric consistency, semantic consistency should be satisfied between the adjacent frames. 
The semantic segmentation of $I_t$ is denoted as $S_t$. According to equation (\ref{equ:projected}), 
the semantic segmentation $\hat{S}_{t' \to t}$ of the $I_t$ frame can be synthesized from $S_{t'}$. 
Different from differentiable bilinear interpolation mechanism in image reconstruction, 
nearest neighbor interpolation is used in semantic segmentation synthesis, 
because the value of semantic segmentation represents the class of each pixel. 
The semantic segmentation reconstruction loss is as follows:
\begin{equation}   
   L_{ss}=\sum_{u,v} \min_{t'} \left[ S_t^{u,v} \neq \hat{S}_{t' \to t}^{u,v} \right] 
\label{equ:ssloss}
\end{equation}
where $\left[ \cdot \right]$is the Iverson bracket.

The projection process in equation (\ref{equ:projected}) implies an assumption: the scene is static and there is no occlusion 
between the two views, but the actual scene obviously does not meet this assumption. 
The projection rule will make mistakes at the pixels of dynamic or occluded objects. 
Pixels that do not obey the semantic consistency may be dynamic objects or occlusion, whose pixels should be removed when 
calculating the reconstruction loss. A mask $M_{t'}$, in which the value of dynamic or occluded pixels is 1 and the value of the remaining pixels is 0, 
is used to indicate these pixels:
\begin{equation}   
  M_{t'}^{u,v} = \left[ S_t^{u,v} \neq \hat{S}_{t' \to t}^{u,v} \right]
\label{equ:mt}
\end{equation}
where $\left[ \cdot \right]$is the Iverson bracket.

The improved image reconstruction loss is:
\begin{equation}   
  L_{img}=\sum_{u,v}\left[ \min_{t'}mre(I_t^{u,v}, \hat{I}_{t' \to t}^{u,v}) < \min_{t'}re(I_t^{u,v}, I_{t'}^{u,v}) \right]\min_{t'}mre(I_t^{u,v}, \hat{I}_{t' \to t}^{u,v})
\label{equ:imgloss}
\end{equation}
where $mre(I_t^{u,v}, \hat{I}_{t' \to t}^{u,v}) = re(I_t^{u,v}, \hat{I}_{t' \to t}^{u,v})+bM_{t'}^{u,v}$ 
and $b$ is a large constant greater than all possible $re(I_t^{u,v}, I_{t'}^{u,v})$ values.

For the depth of some objects, we can give some prior knowledge constraints. 
For roads and sidewalks in autonomous driving datasets, between the two adjacent upper and lower pixels in a image, 
the upper pixel has a longer distance and a larger depth. 
Therefore, we can introduce the following loss:
\begin{equation}   
   L_{road}=\sum_{u,v}R^{u,v}R^{u,v-1}\left[D_t^{u,v}>D_t^{u,v-1}\right]
\label{equ:roadloss}
\end{equation}
where $R^{u,v}$ is 1 if pixel $(u, v)$ is roads or sidewalks, otherwise 0.

\subsection{3D Point Loss}
\label{sec:3dloss}
The photometric loss is mainly concerned with the 2D pixel coordinate system. The 
spatial structure information of 3D points can be used as an effective supervisory 
signal to improve the performance of depth prediction. We propose 3D point loss 
to make full use of 3D information. 
\cite{mahjourian2018unsupervised} aligns 3D point clouds with ICP(Iterative Closest Point algorithm. 
ICP algorithm needs many iterations in the process of point cloud registration, which results in time-consuming network training. 
We use two frame depth maps and transformation matrix to calculate the 3D coordinates 
of corresponding points.

The depth map prediction of $I_t$ can be obtained from depth prediction network, 
and 3D coordinates of each pixel in $I_t$ in the camera coordinate system can be 
further obtained. For a pixel $p_t$ whose coordinates are $(u, v)$ in $I_t$, denote $P_t$ as the coordinates of 3D 
point corresponding to $p_t$ in $I_t$ frame camera coordinate system. $P_t$ can be 
expressed as:
\begin{equation}   
   P_t \sim \hat{D}_t(p_t)K^{-1}p_t.
\label{equ:Pt}
\end{equation}
Pixel coordinates of the corresponding point $p_{t'}$ in $I_{t'}$ can be obtained 
from equation (\ref{equ:projected}).

Denote $\hat{D}_{t'}$ as the depth prediction of $I_{t'}$ by the network. 
The coordinates of $p_{t'}$ are not integers, so the depth value of $p_{t'}$ can't be obtained 
directly. Similar to the image reconstruction, we use bilinear interpolation to estimate the 
depth value of $p_{t'}$. Denote $P_{t'}$ as the coordinates of 3D point corresponding to $p_{t'}$ 
in $I_{t'}$ frame camera coordinate system. Same as equation (\ref{equ:Pt}), $P_{t'}$ can be expressed as:
\begin{equation}   
   P_{t'} \sim \hat{D}_{t'}(p_{t'})K^{-1}p_{t'}.
\end{equation}
Note that $P_t$ and $P_{t'}$ are in different camera coordinates. They need to be transformed 
into the same coordinate system. 

Transform $P_{t'}$ to $I_t$ frame camera coordinate system:
\begin{equation}   
   \hat{P}_{t' \to t} \sim \hat{T}_{t' \to t} P_{t'}
\end{equation}
where $\hat{P}_t$ is the coordinate after transformation, $\hat{T}_{t' \to t}$ is the camera 
coordinate transformation matrix from $I_{t'}$ frame to $I_t$ frame.

\begin{figure}[t]
\begin{center}
\includegraphics[width=0.6\linewidth]{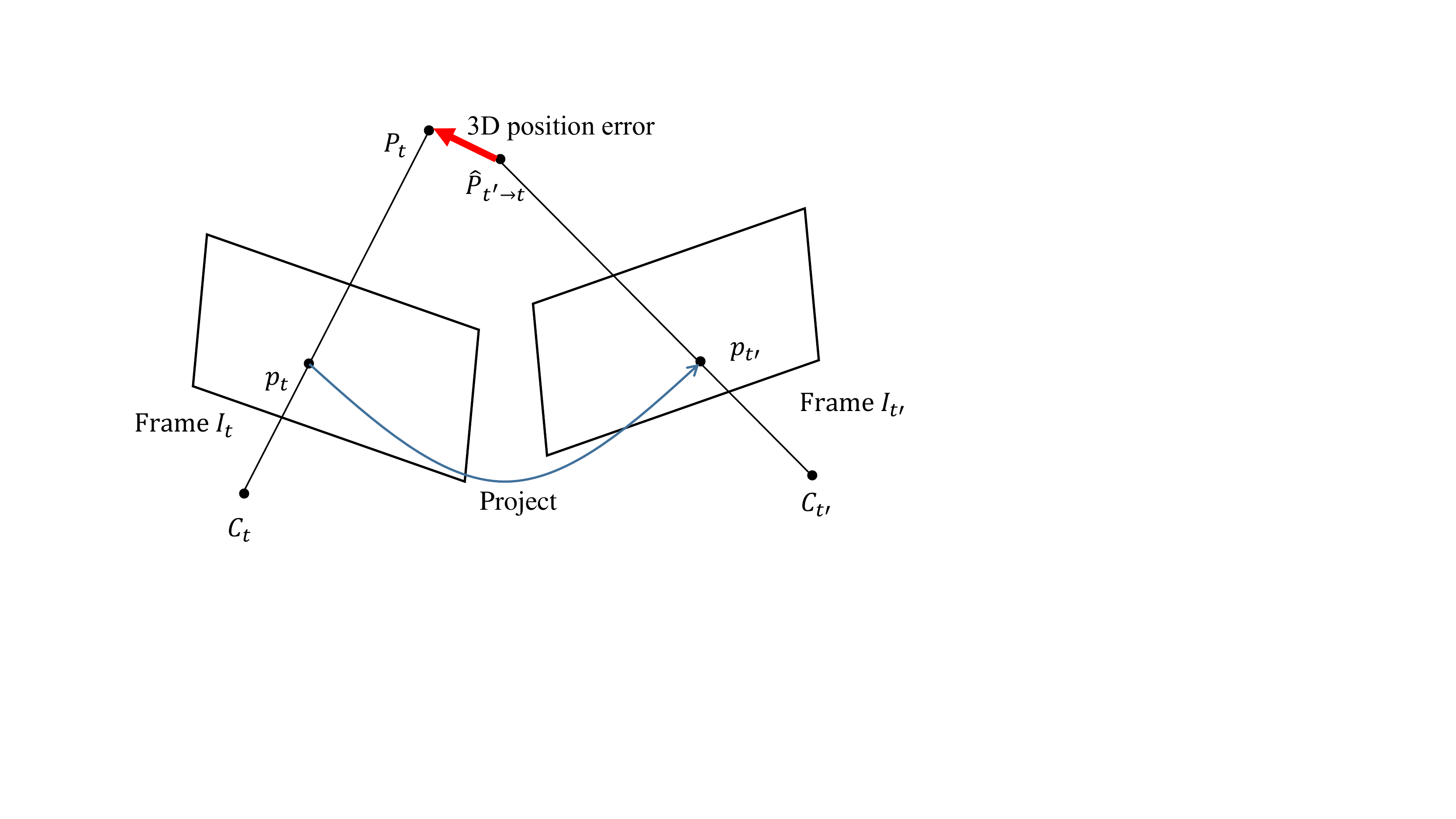}
\end{center}
   \caption{3D point loss. $C_t$ and $C_{t'}$ are camera centers of frame $I_t$ and $I_{t'}$, 
   respectively. For each pixel $p_t$ in frame $I_t$, we can obtaine corresponding pixel $p_{t'}$ in 
   $I_{t'}$ based on the predicted depth map and camera pose. $P_t$ is the corresponding 3D point obtained 
   from the depth map. By interpolating frame $I_{t'}$ depth map, a 3D point corresponding to $p_{t'}$ can be obtained.
   This 3D point is transformed into the coordinate system which point $P_t$ belong to in order to obtain point $\hat{P}_{t' \to t}$. 
   $P_t$ and $\hat{P}_{t' \to t}$ are 3D corresponding points 
   and should be as close as possible}
\label{fig:fig3}
\end{figure}

As shown in Figure~\ref{fig:fig3}, $P_t$ and $\hat{P}_{t' \to t}$ are corresponding points and should be as close as possible. 
The 3D position error is: $pe(P_t, \hat{P}_{t' \to t}) = \Vert P_t - \hat{P}_{t' \to t}\Vert _1$.
Occluded or dynamic pixels should be ignored and the 3D point loss can be expressed as:
\begin{equation}   
  L_{3D}=\sum_{u,v} \left[ \min_{t'}mpe(P_t, \hat{P}_{t' \to t}) < h \right]\min_{t'}mpe(P_t, \hat{P}_{t' \to t})
\end{equation}
where $mpe(P_t, \hat{P}_{t' \to t}) = pe(P_t, \hat{P}_{t' \to t})+hM_{t'}^{u,v}$ 
and $h$ is a large constant greater than all possible $pe(P_t, \hat{P}_{t' \to t})$ values.

Compared with the 2D loss using only one frame depth map, the 3D loss using two depth maps 
based on 3D point consistency, can make better use of 3D spatial structure information.

\subsection{Network Architecture}

The framework is divided into two parts: depth prediction network and pose estimation network. 
Input of the networks includes RGB images and semantic segmentation. 
We adopt the pre-trained semantic segmentation network in \cite{zhu2019improving}, 
which performs fine-tuning on the 200 training images of KITTI dataset \cite{geiger2012we} and can output 19 classes, 
including road, sidewalk, building, wall, fence, etc.
Input each image into the network in \cite{zhu2019improving} to get the result of semantic segmentation. 
The introduction of semantic segmentation network is similar to network pre-training. We use noisy semantic segmentation results, which avoids labeling cost.

Depth prediction network is composed of encoder and decoder networks with skip connections similar to DispNet architecture \cite{mayer2016large}. 
Two parallel encoder networks are used to input a single image and semantic segmentation respectively to extract their feature maps. 
The two feature maps are concatenated and input to the decoder network. 
Each encoder network has 14 convolution layers and kernel size is 3 for all layers, except the first 4 layers for which the sizes are 5, 5, 7, 7 respectively. 
The semantic segmentation, which includes 19 classes, inputs the encoder network in the form of 19 channels. 
The decoder network uses skip-connections to fuse low-level features from different stages of the encoder networks consisting of 7 convolution layers and 7 deconvolution layers.

Pose estimation network takes as input a sequence of adjacent frames concatenated along the color channels, which is similar to the Pose network in \cite{zhou2017unsupervised}.
Different from the relative poses between the target view and each of the source views 
in \cite{zhou2017unsupervised}, the relative poses between every two adjacent frames is predicted.

The total loss function is:
\begin{equation}
\begin{aligned} 
   L=&\lambda_1 L_{img} + \lambda_2 L_{ss} + \lambda_3 L_{3D} + \lambda_4 L_{road} + \lambda_5 L_{smooth}
\end{aligned}
\end{equation}
where $\lambda_1$, $\lambda_2$, $\lambda_3$, $\lambda_4$ and $\lambda_5$ are weights for the different losses. 
Through experiments, we find that setting the weights to $\lambda_1=1$, $\lambda_2=0.1$,
$\lambda_3=0.1$, $\lambda_4=0.1$ and $\lambda_5=0.001$ makes training more stable.

\section{Experiments}
\label{sec:exp}

In this section, we evaluate the performance of our algorithm. We compared our algorithm with prior 
art on both single view depth and pose estimation on KITTI dataset \cite{geiger2012we}. We perform a detailed ablation 
study to show that both the semantic loss and 3D point loss can improve 
the depth prediction and pose estimation performance.

We use the TensorFlow \cite{abadi2016tensorflow} framework to implement the neural network. Adam optimizer 
is used to train the network, with parameters $\beta_1=0.9$, $\beta_2=0.99$, $learning\_rate=0.0002$, 
and $batch\_size=4$. During training, we resize the image sequences to a resolution of $256 \times 832$ which is the same as \cite{luo2018pixel} and \cite{ranjan2019competitive}. 
The network was trained for 10-20 epochs using 3-frame training sequences. The network was trained and 
tested on a NVIDIA GeForce GTX 1080 Ti GPU. The network training time for 200$K$ iterations is about 43 hours. 
The mean inference time of depth map prediction for a image 
of size $256 \times 832$ is 13.6 ms, and the mean inference time of pose estimation for a 3-frame sequence is 5.6 ms.

\subsection{Dataset}
\label{sub:dataset}

We train and evaluate the proposed method on commonly used KITTI benchmark dataset \cite{geiger2012we}, 
which includes a full set of input sources including raw images, 3D point cloud data from LIDAR and camera 
trajectories. We use monocular image sequences for training and test, 3D point cloud and camera trajectories 
are only used to evaluate training models. The original image size is $375 \times 1242$, and images are 
downsampled to $256 \times 832$ during training. In order to compare fairly with other methods, we use two 
different splits of the KITTI dataset to evaluate depth prediction and pose estimation respectively.

We evaluate the single-view depth estimation performance on the test split composed of 697 images from 
28 scenes as in \cite{eigen2014depth}. About 40,000 pictures of the remaining 33 scenes were used for 
training and validation.

The KITTI odometry benchmark \cite{geiger2012we} consists of 22 stereo sequences, 11 sequences (00-10) 
with ground truth trajectories and 11 sequences (11-21) without ground truth. We follow \cite{zhou2017unsupervised} 
to split the KITTI odometry dataset. We train the model on KITTI odometry sequence 00-08 and evaluate 
the pose error on sequence 09 and 10. 

\subsection{Depth Prediction Evaluation}

We evaluate the performance of our monocular depth prediction. The Velodyne laser scanning point is 
projected into the image plane to obtain the ground truth. Since we use only monocular image for training, 
absolute scale information can't be recovered. We multiply the predicted depth map by a scale factor, 
which is the ratio of the median of ground truth to the median of the predicted depth map, the same as \cite{zhou2017unsupervised}. 
Our depth estimation results are compared quantitatively with previous works (some of which use certain type of supervision information). 
All methods are evaluated on the same training images and test images. The split of dataset is 
described in section \ref{sub:dataset}. The error measurements are consistent with those used in \cite{eigen2014depth}.

\setlength{\tabcolsep}{4pt}
\begin{table}
\small
\begin{center}
\caption{Depth evaluation metrics on the KITTI dataset \cite{geiger2012we} using the split of Eigen et al. \cite{eigen2014depth}. For fair comparison, we use the monocular video self-supervision result without pretraining for \cite{godard2019digging}. We mark the best results in bold}
\label{tab:tab1}
\resizebox{\textwidth}{!}{
\begin{tabular}{|l|c|c|c|c|c|c|c|c|c|}
   \hline
   Method                                               & Supervision & Abs Rel        & Sq Rel         & RMSE           & RMSE log       & $\delta<1.25$    & $\delta<1.25^2$   & $\delta<1.25^3$   \\ \hline
   Eigen \cite{eigen2014depth} Coarse              & Depth       & 0.214          & 1.605          & 6.563          & 0.292          & 0.638            & 0.804             & 0.894             \\
   Eigen \cite{eigen2014depth} Fine                & Depth       & 0.203          & 1.548          & 6.307          & 0.282          & 0.702            & 0.890             & 0.958             \\
   Liu \cite{liu2015learning}                      & Depth       & 0.201          & 1.584          & 6.471          & 0.273          & 0.680            & 0.898             & 0.967             \\ 
   Zhou \cite{zhou2017unsupervised}                & No          & 0.208          & 1.768          & 6.856          & 0.283          & 0.678            & 0.885             & 0.957             \\
   Mahjourian \cite{mahjourian2018unsupervised}    & No          & 0.163          & 1.240          & 6.220          & 0.250          & 0.762            & 0.916             & 0.968             \\
   Yang \cite{yang2018lego}                        & No          & 0.162          & 1.352          & 6.276          & 0.252          & 0.783            & 0.921             & 0.969             \\
   Yin \cite{yin2018geonet}                        & No          & 0.155          & 1.296          & 5.857          & 0.233          & 0.793            & 0.931             & 0.973             \\
   Luo \cite{luo2018pixel}                         & No          & 0.141          & 1.029          & 5.350          & 0.216          & 0.816            & 0.941             & 0.976             \\
   Ranjan \cite{ranjan2019competitive}             & No          & 0.140          & 1.070          & 5.326          & 0.217          & 0.826            & 0.941             & 0.975             \\
   Godard \cite{godard2019digging}                 & No          & 0.132          & 1.044          & 5.142          & 0.210          & \textbf{0.845}   & 0.948             & 0.977             \\ \hline
   Ours                                            & No          & \textbf{0.131} & \textbf{0.902} & \textbf{4.980}  & \textbf{0.204} & 0.837            & \textbf{0.952}    &  \textbf{0.981}   \\ \hline 
  
\end{tabular}}
\end{center}
\end{table}
\setlength{\tabcolsep}{1.4pt}

\begin{figure*}
   \begin{center} 
   \includegraphics[width=1\linewidth]{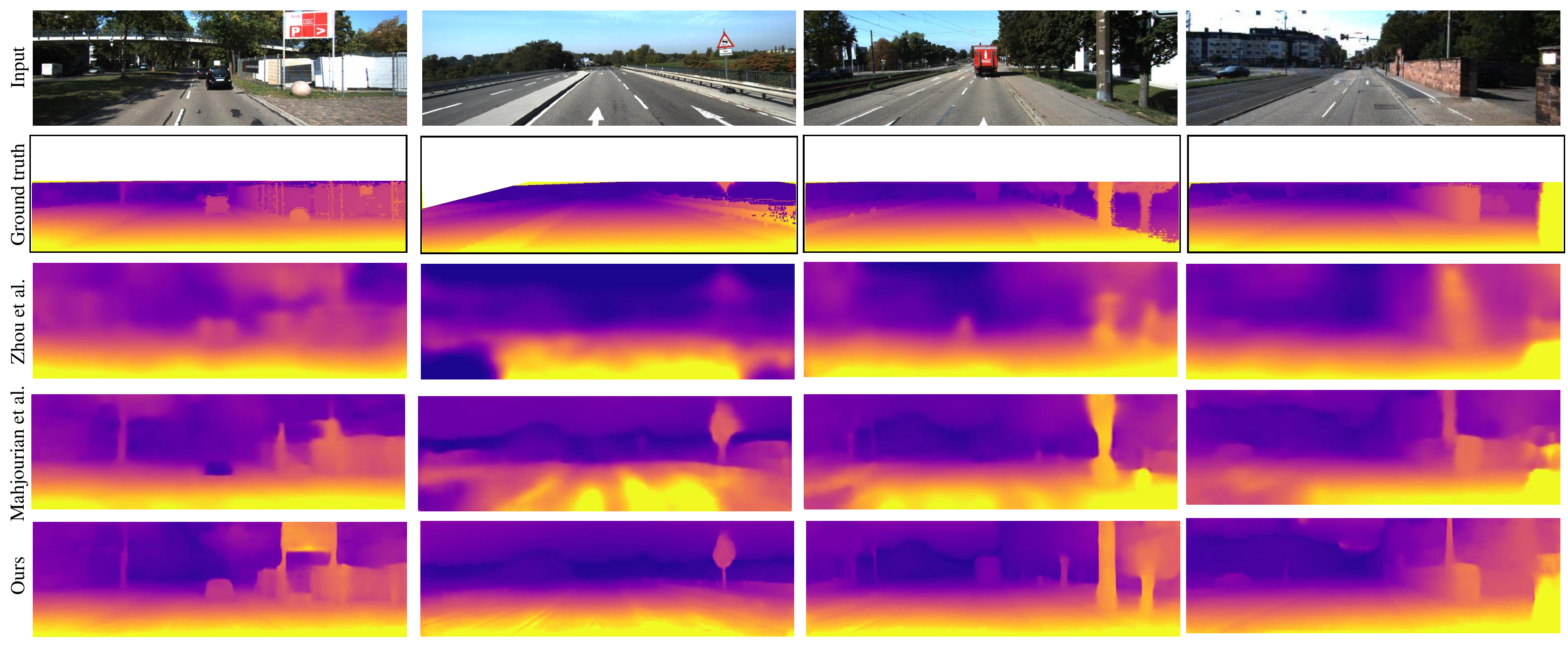}
   \end{center}
      \caption{Some examples of comparison with Zhou et al. \cite{zhou2017unsupervised},
      Mahjourian et al. \cite{mahjourian2018unsupervised} and ours on KITTI dataset \cite{geiger2012we}. 
      The sparse ground truth depth is interpolated from LIDAR for visualization purpose. Our results have clearer object boundaries and 
      better depth prediction, especially for trees and cars}
   \label{fig:fig4}
   \end{figure*}

Table~\ref{tab:tab1} shows the comparison between our method and other methods. Abs Rel, Sq Rel, RMSE and RMSE 
log are error metrics, and small values mean better performance. $\delta<1.25$, $\delta<1.25^2$ and $\delta<1.25^3$ 
are accuracy metrics, and large values mean better performance. In order to make a fair comparison with other methods, 
we use the maximum depth thresholds of 80 meters to evaluate. 
Our method achieves the best performance on most metrics.

Figure~\ref{fig:fig4} shows some visualization examples compared with other methods. As can be seen from Figure~\ref{fig:fig4}, 
the depth maps predicted by our method are clearer at the boundary of objects and better recover the depth of 
cars and trees. 

\setlength{\tabcolsep}{4pt}
\begin{table}
\begin{center}
\caption{Absolute Trajectory Error (ATE) on KITTI odometry dataset \cite{geiger2012we} over all multi-frame snippets}
\label{tab:tab2}
\begin{tabular}{|l|c|c|}
   \hline
   Method                                             & Seq. 09                  & Seq. 10              \\ \hline
   ORB-SLAM (full)                                    & $0.014 \pm 0.008$ m       & $0.012 \pm 0.011$ m    \\ 
   ORB-SLAM (short)                                   & $0.064 \pm 0.141$ m        & $0.064 \pm 0.130$ m    \\ 
   Mean Odom.                                         & $0.032 \pm 0.026$ m        & $0.028 \pm 0.023$ m    \\ 
   Zhou \cite{zhou2017unsupervised}             & $0.021 \pm 0.017$ m        & $0.020 \pm 0.015$ m    \\ 
   Mahjourian \cite{mahjourian2018unsupervised} & $0.013 \pm 0.010$ m        & $0.012 \pm 0.011$ m    \\ 
   Yin \cite{yin2018geonet}                     & $0.012 \pm 0.007$ m        & $0.012 \pm 0.009$ m    \\ 
   Luo \cite{luo2018pixel}                      & $0.013 \pm 0.007$ m        & $0.012 \pm 0.008$ m    \\ 
   Ranjan \cite{ranjan2019competitive}          & $0.012 \pm 0.007$ m        & $0.012 \pm 0.008$ m    \\
   Godard \cite{godard2019digging}              & $0.017 \pm 0.008$ m        & $0.015 \pm 0.010$ m    \\ \hline
   Ours                                               & $\mathbf{0.010 \pm 0.005}$ m& $\mathbf{0.009 \pm 0.008}$ m    \\ \hline
\end{tabular}
\end{center}
\end{table}
\setlength{\tabcolsep}{1.4pt}
\subsection{Pose Estimation Evaluation}

We use KITTI odometry dataset to evaluate the proposed approach, and compare the results with Zhou 
et al. \cite{zhou2017unsupervised}, Mahjourian et al. \cite{mahjourian2018unsupervised}, Yin et al. \cite{yin2018geonet}, Luo et al. \cite{luo2018pixel}, 
Godard et al. \cite{godard2019digging}, Ranjan et al. \cite{ranjan2019competitive} and 
ORB-SLAM \cite{mur2015orb} proposed by Mur-Artal et al.. We also use the dataset mean of car motion (using ground truth odometry) 
for 5-frame snippets as another baseline for comparison. Among them, \cite{zhou2017unsupervised,mahjourian2018unsupervised,yin2018geonet,luo2018pixel,ranjan2019competitive,godard2019digging} are unsupervised deep learning methods, while 
ORB-SLAM is a traditional geometry-based method. The methods based on deep learning use the same 
training data. The models are trained on the KITTI odometry dataset 00-08 sequences and the relative 
pose estimation is evaluated on the sequences 09 and 10. In the experiment, we fixed the length 
of the input image sequences to 3 frames, which is the same as \cite{mahjourian2018unsupervised}. 
We compared two versions of ORB-SLAM. ``ORB-SLAM (full)" accepts all frames of the whole 
sequence as input, which involves global optimization steps, such as loop closure detection and 
bundle adjustment. Note that there is no loop in sequence 10, so loop closure detection is not used. 
``ORB-SLAM (short)" only accepts five consecutive frames as input. Because of the scale uncertainty 
of monocular VO, we optimize the scale to make the trajectory consistent with the ground truth.

In order to make a fair comparison with other methods, like \cite{zhou2017unsupervised,mahjourian2018unsupervised,yin2018geonet,luo2018pixel,ranjan2019competitive}, we measure the Absolute Trajectory Error (ATE) \cite{sturm2012benchmark} over 3 or 5 
frame snippets as the metric for pose evaluation. 
As shown in Table~\ref{tab:tab2}, our method is superior to other methods. The output of pose estimation network is the relative poses between 3 or 5 frames snippets. 
Compared with the geometry-based method, the camera trajectory predicted by this kind of method has a larger cumulative error for a long time image sequence.

\subsection{Ablation Study}

We investigate the contribution of several components proposed in our unsupervised architecture. 
As shown in Table~\ref{tab:tab3}, in order to demonstrate the importance of each component 
of the losses, we conducted ablation studies on depth prediction. We trained and evaluated 
three models with different losses. The experimental results show the 
importance of each component.

\setlength{\tabcolsep}{4pt}
\begin{table}
\small
\begin{center}
\caption{Depth evaluation metrics on the KITTI dataset \cite{geiger2012we} using the split of Eigen et al. \cite{eigen2014depth} for various versions of our model}
\label{tab:tab3}
\resizebox{\textwidth}{!}{
\begin{tabular}{|cccc|c|c|c|c|c|c|c|}
   \hline
   $L_{img}$   & $L_{ss}$   & $L_{road}$   & $L_{3D}$  & Abs Rel        & Sq Rel         & RMSE           & RMSE log       & $\delta<1.25$    & $\delta<1.25^2$   & $\delta<1.25^3$   \\ \hline
   $\surd$     &            &              &           & 0.144          & 1.089          & 5.423          & 0.214          & 0.815            & 0.945             & 0.979             \\
   $\surd$     & $\surd$    &              &           & 0.138          & 1.007          & 5.235          & 0.211          & 0.829            & 0.948             & 0.979             \\
   $\surd$     & $\surd$    & $\surd$      &           & 0.133          & 0.939          & 5.157          & 0.208          & \textbf{0.837}            & 0.951             & 0.980             \\
   $\surd$     &            &              & $\surd$   & 0.136          & 0.913          & 5.191          & 0.210          & 0.820            & 0.947             & \textbf{0.981}             \\ \hline
   $\surd$     & $\surd$    & $\surd$      & $\surd$   & \textbf{0.131} & \textbf{0.902} & \textbf{4.980}  & \textbf{0.204} &\textbf{0.837}   & \textbf{0.952}    &  \textbf{0.981}    \\
   \hline 
  
\end{tabular}}
\end{center}
\end{table}
\setlength{\tabcolsep}{1.4pt}

\begin{figure}
   \begin{center}
   \includegraphics[width=1\linewidth]{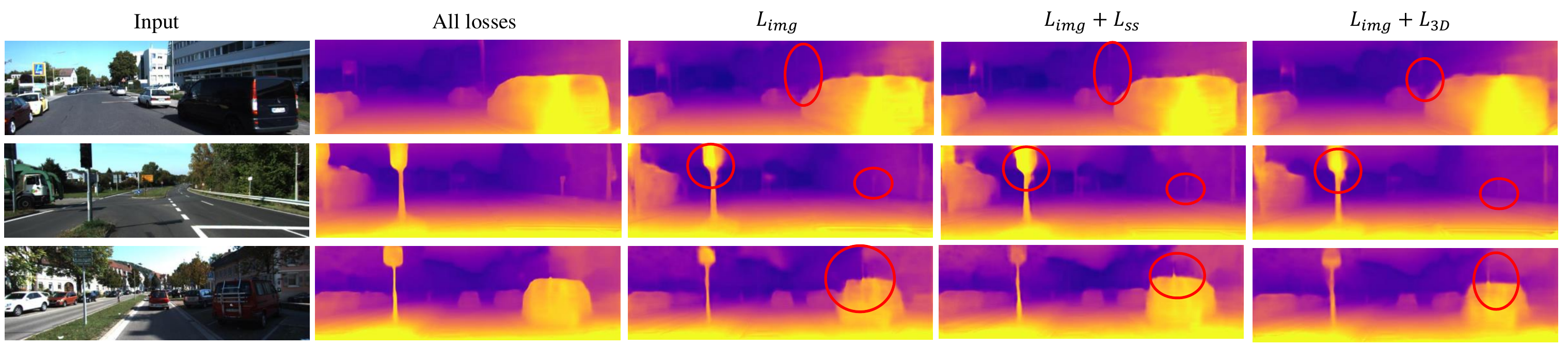}
   \end{center}
      \caption{Examples of depth estimation results under different loss training. The failure part of depth prediction is marked with red circles box}
   \label{fig:fig5}
\end{figure}

\setlength{\tabcolsep}{4pt}
\begin{table}
\begin{center}
   \begin{tabular}{|cccc|c|c|}
      \hline
      $L_{img}$   & $L_{ss}$   & $L_{road}$   & $L_{3D}$    & Seq. 09                  & Seq. 10              \\ \hline
      $\surd$     &            &              &             &  $0.013 \pm 0.009$ m        & $0.012 \pm 0.009$ m   \\   
      $\surd$     & $\surd$    &  $\surd$     &             &  $0.010 \pm 0.006$ m                      & $0.010 \pm 0.009$ m   \\ 
      $\surd$     &            &              & $\surd$     & $0.010 \pm 0.006$ m             &$0.010 \pm 0.009$ m    \\  \hline
      $\surd$     & $\surd$     &  $\surd$    & $\surd$     & $\mathbf{0.010 \pm 0.005}$ m               & $\mathbf{0.009 \pm 0.008}$ m      \\  \hline
      \end{tabular}
\end{center}
\caption{Absolute Trajectory Error (ATE) on KITTI odometry dataset \cite{geiger2012we} for various versions of our model}
\label{tab:tab4}
\end{table}
\setlength{\tabcolsep}{1.4pt}

Figure~\ref{fig:fig5} shows the depth maps generated by the models under different loss function 
training. Using all losses can get the best performance. The boundary of objects is clearer, and it is better to predict the depth of small or thin objects such as poles and traffic signs. 

As shown in Table~\ref{tab:tab4}, we compare the effects of different loss functions on pose estimation results. 
We can see that semantic loss and 3D point loss improve the accuracy of pose estimation, although not so significant compared to depth estimation. 

\section{Conclusions}
\label{sec:con}

We propose a semantics-driven unsupervised deep learning method for monocular depth prediction and camera 
ego-motion estimation tasks. It is trained on unlabeled monocular image sequence, and performs 
pose estimation and dense depth map estimation during testing. 
We introduce a semantic loss to reduce the impact of dynamic objects or occluded objects in the scene and improve depth estimation performance by considering the semantic consistency and correlation between depth and semantics. 
We also propose a new 3D point loss to improve the accuracy of depth prediction. The experimental evaluation on the KITTI dataset shows 
that our method achieves good good performance. 
Compared with semantic segmentation, the boundary between objects in depth map is not clear enough. 
One direction of future work is to use semantic segmentation to improve the boundary performance between objects in depth maps.

%
%
\bibliographystyle{splncs04}
\bibliography{depthbib}
\end{document}